\documentclass[runningheads]{llncs}
\usepackage[T1]{fontenc}
\usepackage{graphicx,verbatim}

\usepackage{amsmath}
\usepackage{amsfonts}
\newcommand{\blue}{\textcolor{black}}
\newcommand{\red}{\textcolor{black}}
\newcommand{\reddd}{\textcolor{red}}
\newcommand{\blueee}{\textcolor{blue}}
\usepackage{xcolor}
\usepackage{booktabs}
\usepackage{multirow}

\begin{document}
\title{Precision Recall Controllable Radiology Report Generation via Hybrid Natural Language and Clinical Reward Learning}
\titlerunning{Precision-Recall Controllable Radiology Report Generation}
%

\author{\blue{
Ling Chen\inst{1} \and
Ruinan Jin\inst{1} \and
Jun Luo\inst{1} \and
Hanliang Chen\inst{1} \and
Quirin Strotzer\inst{2} \and
Rongkai Yan\inst{1} \and
Yuan Xue\inst{1} \and
Luciano Prevedello\inst{1} \and
Dufan Wu\inst{1}}
}


\authorrunning{\blue{L. Chen et al.}}

\institute{
\blue{The Ohio State University, Columbus, OH 43221, USA} \\
\and
\blue{University Medical Center Regensburg, Regensburg, Germany} \\
\email{\blue{lingchen.chen@osumc.edu, dufan.wu@osumc.edu}}
}

\maketitle              
\begin{abstract}
Automated radiology report generation (RRG) has gained increasing attention because it can reduce the heavy workload of clinical report writing. However, most existing methods mainly optimize for natural language generation (NLG) metrics that focus on language fluency, while providing little control over clinically important factors such as precision and recall. As consequence, generated reports may be fluent but not well aligned with different clinical needs.
To address this challenge, we propose a reinforcement learning framework for precision recall controllable RRG, where a control parameter explicitly adjusts the trade-off between clinical precision and recall during inference. This design allows the model to flexibly generate reports according to different clinical requirements. To ensure clinical correctness, we introduce a \blue{clinical reward}  into the training objective, which helps improve clinical efficacy (CE) beyond standard language-based optimization. In addition, we apply a group-relative training strategy that normalizes rewards within each training group, reducing reward variance and improving training stability.
Extensive experiments on the MIMIC-CXR dataset show that our method consistently outperforms state-of-the-art approaches in both NLG{ and CE} evaluation metrics, while providing reliable control over the CE precision recall trade-off.

\keywords{Radiology Report Generation \and Receiver Operation Curve \and Reinforcement Learning  \and Group-Relative Training \and Chest X-ray.}

\end{abstract}

\section{Introduction}

Automated radiology report generation (RRG) aims to produce free-text clinical reports directly from medical images, such as chest X-rays, and has received growing attention in recent years. By reducing the time and effort required for manual report writing, these systems have the potential to improve clinical workflow efficiency and support large-scale medical imaging analysis. Most existing approaches treat report generation as a sequence-to-sequence learning problem and are commonly trained using maximum likelihood estimation (MLE) with word-level supervision \cite{jing2018automatic,chen2020generating}.

While MLE-based methods can generate fluent and grammatically correct reports, they often suffer from exposure bias and show limited alignment with clinical goals \cite{yu2023evaluating} \cite{sloan2024automated}. To address these issues, recent studies have introduced reinforcement learning (RL)–based optimization strategies that directly optimize sentence-level evaluation metrics \cite{zhang2020optimizing,li2018hybrid,yi2025lhr,qin2022reinforced,celikyilmaz2018deep}. However, most prior work focuses on improving natural language generation (NLG) scores, such as BLEU \cite{papineni2002bleu}, METEOR \cite{banerjee2005meteor}, and ROUGE-L \cite{lin2004rouge}, which mainly measure text similarity and linguistic quality. As a result, generated reports may appear fluent but still contain clinically incorrect, incomplete, or unbalanced findings.
More importantly, existing methods provide little control over clinically critical trade-offs during report generation. In real-world practice, different clinical scenarios may require different emphasis on precision and recall \cite{topol2019high}. For example, screening exams often favor high recall to avoid missing abnormalities, while diagnostic tests may prioritize high precision to reduce false positives. Current report generation models lack explicit mechanisms to adjust such trade-offs, limiting their practical usability in diverse clinical settings. 

In this work, we propose a reinforcement learning framework for precision recall controllable RRG for chest X-rays, in which a continuous control parameter explicitly adjusts the balance between clinical precision and recall at inference time. This design allows the same model to flexibly generate reports that align with different clinical requirements, without retraining or changing decoding strategies. 
To improve clinical performance, we further introduce a \blue{clinical reward}  explicitly supervises the presence and absence of clinically relevant findings, enhancing clinical efficacy beyond language-only optimization.
In addition, reinforcement learning for report generation often suffers from high reward variance and unstable training \cite{paulus2017deep}. To address this challenge, we adopt a group-relative training strategy \cite{shao2024deepseekmath} that normalizes rewards within each sample group, effectively reducing variance and stabilizing policy optimization. This strategy improves learning efficiency and leads to more consistent performance gains across datasets.
We evaluate the proposed method on MIMIC-CXR \cite{johnson2019mimic} dataset. Experimental results show that our approach consistently outperforms state-of-the-art methods in both language quality and clinically oriented evaluation metrics, while providing reliable and interpretable control over the precision recall trade-off.

\section{Method}
\begin{figure}[t]
    \centering
    \includegraphics[width=0.99\linewidth]{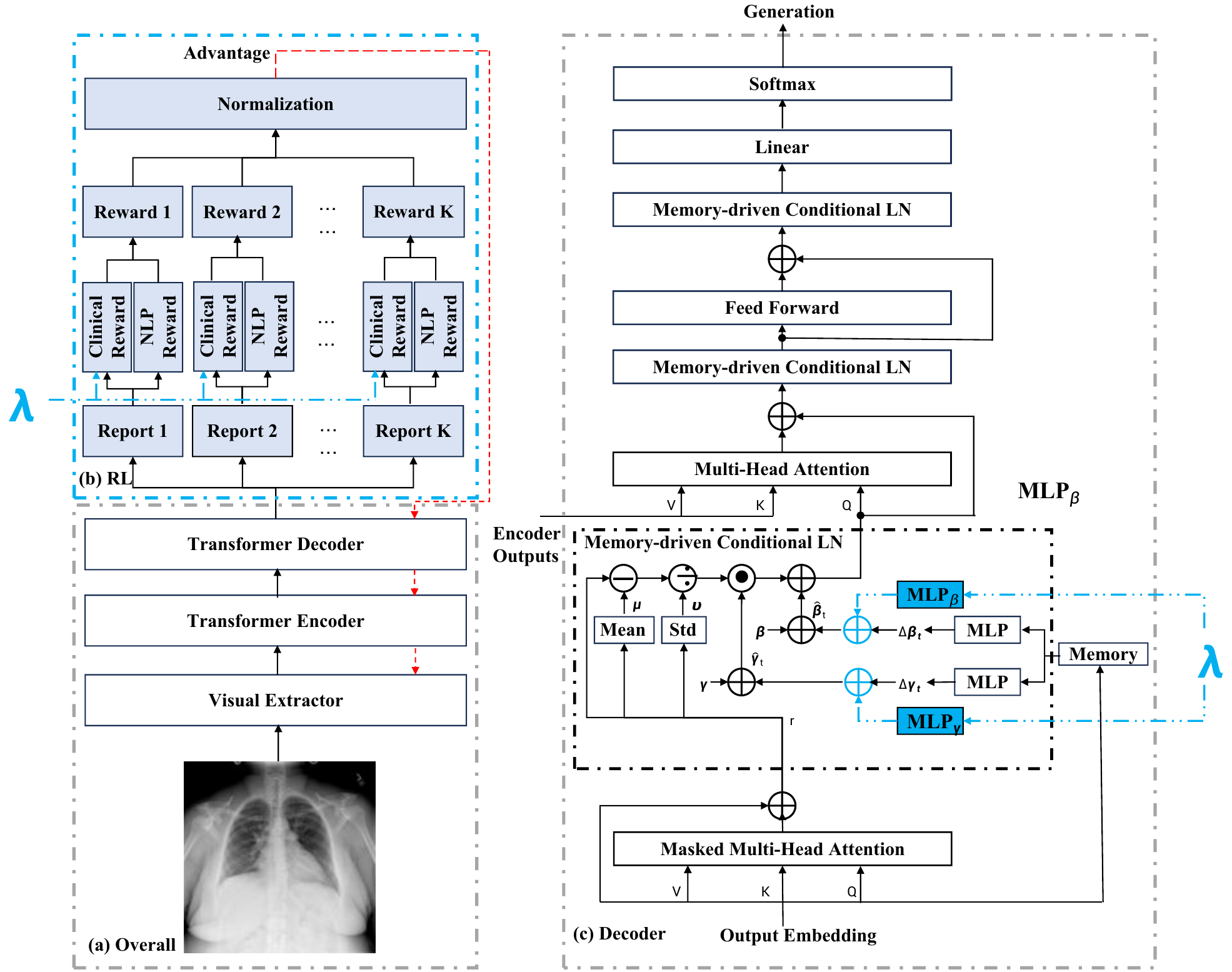}
    \caption{{\blue{Overall framework} of the proposed precision recall controllable RRG model.
(a) Overall architecture. (b) Reinforcement learning process. (c) Transformer decoder. The overall framework and decoder are shown in gray dashed boxes with details omitted. Red dotted lines denote gradient back-propagation during training. Blue dotted lines indicate the precision recall control parameter $\lambda$ applied at both the representation and reward levels. The group-relative training component is highlighted in blue dashed boxes.}}
    \label{fig:overall_framework}
\end{figure}

The overall framework is illustrated in Fig.~\ref{fig:overall_framework}, which consists of the overall architecture (Fig.~\ref{fig:overall_framework}(a)), the reinforcement learning process (Fig.~\ref{fig:overall_framework}(b)), and the transformer encoder details (Fig.~\ref{fig:overall_framework}(c)).

\subsection{Precision Recall Controllable Reinforcement Learning}

We formulate RRG as a reinforcement learning problem, where the model generates a report sequence
$Y = (y_1, \dots, y_T)$ conditioned on an input image $X$.

To explicitly control the precision recall trade-off, we introduce a continuous control parameter
$\lambda \in [0,1]$ that is consistently applied at both the representation level and the reward level.
This unified design ensures coherent controllability throughout the generation and optimization process.

At the representation level, inspired by conditional normalization techniques such as FiLM \cite{perez2018film}, we introduce a Precision Recall Conditioned Adaptive Layer Normalization (PRC-AdaLN) module to incorporate the clinical control parameter $\lambda$ into the decoder representations. Specifically,
\begin{equation}
\text{PRC-AdaLN}(x,\lambda)
=
\big(\gamma + \Delta\gamma + \blue{\text{MLP}_{\gamma}}(\lambda)\big)
\odot
\frac{x - \mu(x)}{\sigma(x)}
+
\big(\beta + \Delta\beta + \blue{\text{MLP}_{\beta}}(\lambda)\big),
\end{equation}
where $x$ denotes the decoder feature, $\mu(x)$ and $\sigma(x)$ are the mean and standard deviation computed along the feature dimension, and $\gamma$ and $\beta$ are the learnable affine parameters in standard LayerNorm. The terms $\Delta\gamma$ and $\Delta\beta$ denote additional residual modulation parameters, \blue{while two lightweight MLPs, $\text{MLP}_{\gamma}$ and $\text{MLP}_{\beta}$, map the clinical control parameter $\lambda$ to separate conditioning offsets for the normalization scale and bias, respectively}. As illustrated in Fig.~\ref{fig:overall_framework}(c), this representation level modulation dynamically adjusts feature statistics, enabling the decoder to explicitly steer generation along the clinically meaningful precision recall axis.


At the reward level, $\lambda$ further controls the optimization objective by balancing
precision- and recall-oriented clinical efficacy (CE) rewards:
\begin{equation}\label{eq:reward_ce}
R_{\mathrm{CE}}(\lambda)
= \lambda \, R_{\mathrm{prec}} + (1 - \lambda) \, R_{\mathrm{rec}} .
\end{equation}

During training, for each training sample, the representation modulation and reward optimization shared a random $\lambda$ sampled uniformly from $[0,1]$.
Thus the model learns a family of conditional policies with consistent precision recall characteristics.
At inference time, adjusting $\lambda$ allows a single trained model to flexibly generate reports
that emphasize either higher precision or higher recall, without retraining or modifying the decoding strategy.

\subsection{Hybrid Natural Language and Clinical Reward Learning}

We optimize the report generation model using a hybrid training objective that combines
reinforcement learning with standard cross-entropy loss:
\begin{equation}
\label{eq:total_loss}
\mathcal{L}_{\text{total}} = \alpha \mathcal{L}_{\text{rl}} + (1 - \alpha)\mathcal{L}_{\text{ce}},
\end{equation}
where $\alpha$ balances reinforcement learning loss and cross-entropy loss.

For the reinforcement learning component, we minimize the negative expected reward:
\begin{equation}\label{eq:rl_loss}
\mathcal{L}_{\text{rl}} = -\mathbb{E}_{Y \sim p_\theta}[R(Y)],
\end{equation}
where $\theta$ refers to the model parameters and the reward function $R(Y)$ integrates both natural language generation (NLG) quality
and CE:
\begin{equation}
\label{eq:total_reward}
R(Y) = \rho R_{\text{NLG}}(Y) + (1 - \rho) R_{\text{CE}}(Y).
\end{equation}

Here, $\rho \in [0,1]$ is a fixed hyperparameter that controls the trade-off between natural language quality and clinical efficacy. 
\blue{The clinical reward is based on the clinical efficacy metric, with $R_{\text{CE}}(Y)$ computed according to Eq.~\ref{eq:reward_ce}.
}
The natural language generation reward $R_{\text{NLG}}(Y)$ is defined as:
\begin{equation}
\label{eq:nlg_reward}
R_{\text{NLG}}(Y) = \sum_{i} \omega_i r_i(Y),
\end{equation}
where $r_i(Y)$ represents multiple NLG evaluation metrics, including BLEU-4, METEOR, and ROUGE-L \cite{yi2025lhr},
and $\omega_i$ are non-negative weighting coefficients with $\sum_i \omega_i = 1$.



\subsection{Group-Relative Training}

Reinforcement learning for report generation often suffers from high reward variance,
which can lead to unstable training, especially when multiple reward components are involved.
To alleviate this issue, we adopt a group-relative self-critical sequence training strategy \cite{shao2024deepseekmath}.

For each input image $x$, we perform group sampling by drawing $K$ candidate reports
$\{Y^{(1)}, Y^{(2)}, \dots, Y^{(K)}\}$ from the current policy.
Each sampled report is evaluated using the task-specific reward function,
yielding a set of rewards $\{R^{(k)}\}_{k=1}^{K}$.

Instead of relying on absolute reward values, we compute a group-relative baseline
by normalizing rewards within each group:
\begin{equation}
\hat{A}^{(k)} =
\frac{R^{(k)} - \mathrm{mean}(\{R^{(i)}\}_{i=1}^{K})}
{\mathrm{std}(\{R^{(i)}\}_{i=1}^{K}) + \epsilon},
\end{equation}
where $\hat{A}^{(k)}$ denotes the standardized advantage of the $k$-th sampled report
and $\epsilon$ is a small constant for numerical stability. The reinforcement learning loss is computed according to Eq.~\ref{eq:rl_loss} by replacing $R(Y)$ with $\hat{A}$ for each sample. We then use the REINFORCE algorithm \cite{williams1992simple} to compute policy gradients and update the model parameters.

By using the group-relative standardized advantage as the learning signal,
the model focuses on the relative quality of candidate reports generated for the same image.
This strategy effectively reduces reward variance and stabilizes policy optimization.

\section{Experiments}
\subsubsection{Datasets and Evaluation Metrics}
We evaluate our model on the MIMIC-CXR dataset \cite{johnson2019mimic}, a large-scale chest X-ray benchmark containing 473,057 radiographs paired with 206,563 reports. Following the preprocessing protocol in \cite{chen2020generating}, we adopt the official patient-level split with 270,790 training, 2,130 validation, and 3,858 testing samples. Each study is paired with its corresponding report as ground truth. We only used the findings section for each report. 

Following prior work \cite{yi2025lhr,liu2025rrg}, we evaluate model performance using a combination of natural language generation (NLG) metrics and { CE metrics. The NLG metrics include BLEU \cite{papineni2002bleu}, METEOR \cite{banerjee2005meteor}, and ROUGE-L \cite{lin2004rouge}. 
\blue{For CE evaluation, generated and ground-truth reports are first converted into 14 CheXbert labels~\cite{smit2020chexbert}. Following our training setup, uncertain findings are considered positive, while negative and blank labels are considered negative. We then report micro-averaged precision, recall, and F1 score across all labels and test samples.}

\subsubsection{Implementation Details}
We used $\alpha = 0.99$ in Eq. \ref{eq:total_loss} that balances RL and cross-entropy loss, $\rho=0.1$ in Eq. \ref{eq:total_reward} that balances NLG and clinical efficacy. Following \cite{yi2025lhr}, the weighting coefficients for the NLG reward in Eq. \ref{eq:nlg_reward} were set to $\omega_1 = 5/11$, $\omega_2 = 5/11$, and $\omega_3 = 1/11$ for BLEU-4, METEOR, and ROUGE-L, respectively. For the group-relative training, we set the group size $K=5$. All remaining hyperparameters are kept consistent with \cite{yi2025lhr}.
A pretrained model \cite{chen2020generating} was fine-tuned on the MIMIC-CXR dataset for 5 epochs with a batch size of 4 using 16 NVIDIA A100 GPUs \blue{provided by the Ohio Supercomputer Center~\cite{Ohio_Supercomputer_Center1987-dl}}. We further applied post-processing to all generated reports to remove repeating sentences and sentences that were shorter than three words.
{Our code is available at: 
\blue{https://github.com/98lingchen/MICCAI2026}.

\section{Results}
\subsubsection{Parameter Performance}

As shown in Fig.~\ref{fig:lambda_ablation}, the precision recall control parameter $\lambda$ provides effective and continuous regulation over clinical precision and recall. We performed inference using different $\lambda$ values from 0 to 1 with an interval of 0.1. As $\lambda$ increases, clinical precision consistently improves while recall gradually decreases, demonstrating a clear trade-off between precision and recall. The F1-score exhibits a unimodal trend and reaches its maximum at {$\lambda = 0.3$}. These results validate that $\lambda$ enables flexible adjustment of report generation behavior to meet different clinical requirements, such as high-recall screening and high-precision confirmation, without requiring model retraining. 
For the results reported below, we use $\lambda = 0.3$, which achieved the best F1 scores.
\begin{figure}[h]
    \centering
    \includegraphics[width=0.5\linewidth]{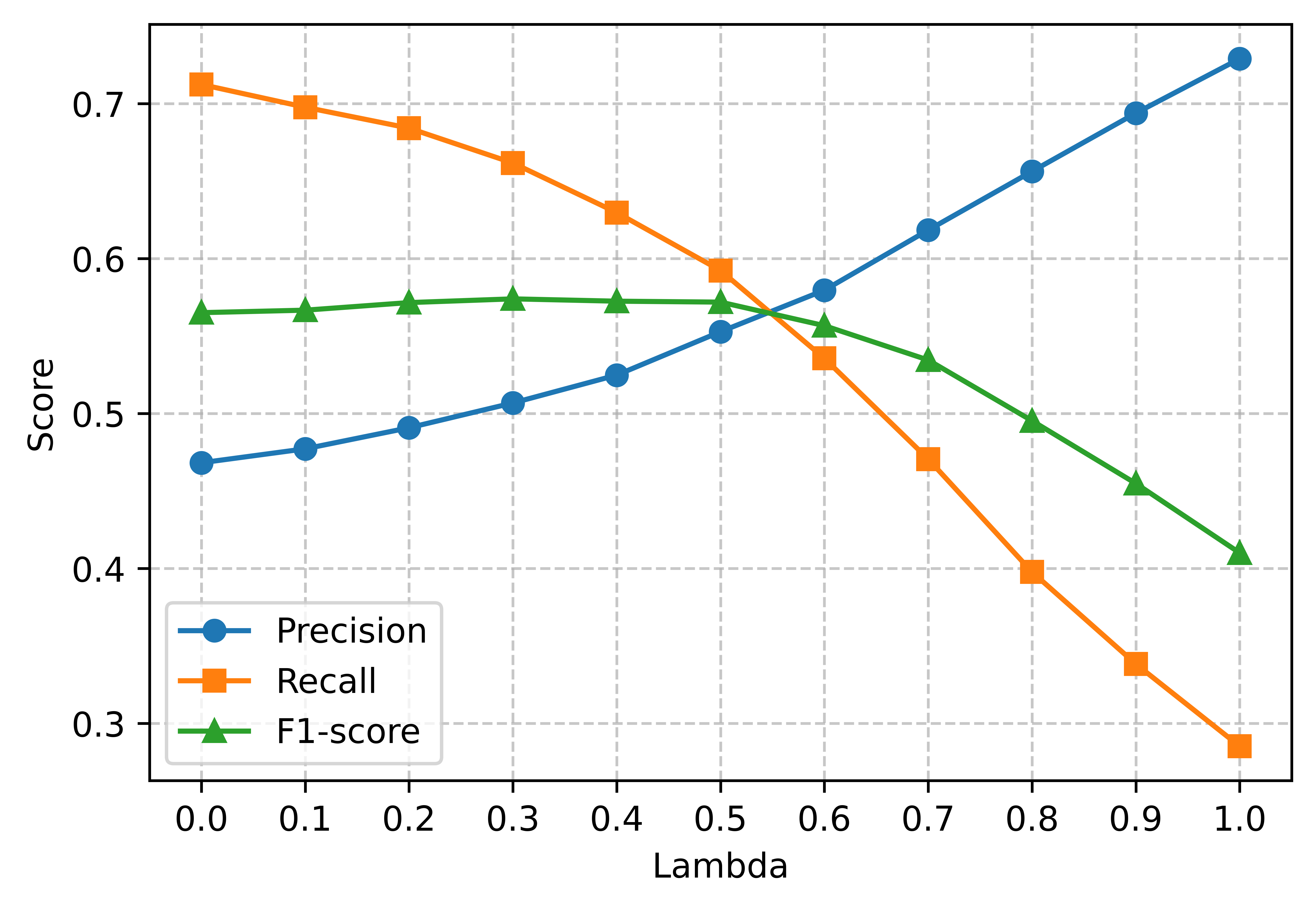}
    \caption{{Effect of the precision recall control parameter $\lambda$ on clinical precision, recall, and F1-score.}}
    \label{fig:lambda_ablation}
\end{figure}
\subsubsection{Performance Comparison}

We compared our method with several state-of-the-art methods on the MIMIC-CXR dataset, including R2Gen \cite{chen2020generating}, METrans \cite{wang2023metransformer}, R2GenGPT \cite{wang2023r2gengpt}, BoostRRG \cite{sha2025contrastive}, Diff-RRG \cite{yun2025diff}, MLRG \cite{liu2025enhanced}, and MedGemma 1.5 4B \cite{sellergren2025medgemma}. 
Results of the methods except MedGemma were reported as provided in their original papers. For MedGemma, the reports were generated as follows:

\begin{table}[h]
\centering
\caption{Performance comparisons between the proposed method and existing approaches on the test set of the MIMIC-CXR dataset using both NLG and CE metrics.}
\label{tab:iu_mimic_results}
\setlength{\tabcolsep}{4.5pt}
\scriptsize
\begin{tabular}{l cccccc ccc}
\toprule
\multirow{2}{*}{Method} 
& \multicolumn{6}{c}{NLG Metrics} 
& \multicolumn{3}{c}{CE Metrics} \\ 
\cmidrule(lr){2-7} \cmidrule(lr){8-10}
& B-1 & B-2 & B-3 & B-4 & RG-L & MET. & PREC. & REC. & F1 \\ 
\midrule

R2Gen       & 0.353 & 0.218 & 0.145 & 0.103 & 0.277 & 0.142 & 0.333 & 0.273 & 0.276 \\
METrans     & 0.386 & 0.250 & 0.169 & 0.124 & 0.291 & 0.152 & 0.364 & 0.309 & 0.334 \\
R2GenGPT    & 0.411 & 0.267 & 0.186 & 0.134 & 0.297 & 0.160 & 0.392 & 0.387 & 0.389 \\
BoostRRG    & 0.402 & 0.262 & 0.180 & 0.128 & 0.291 & {0.175} & 0.465 & 0.482 & 0.473 \\
Diff-RRG    & 0.405 & 0.251 & 0.169 & 0.120 & 0.276 & 0.164 & 0.528 & 0.430 & 0.474 \\
MLRG        & 0.411 & 0.277 & 0.204 & {0.158} & 0.320 & {0.176} & \textbf{0.549} & 0.468 & 0.505 \\
MedGemma    & 0.200 & 0.114 & 0.071 & 0.045 & 0.160 & 0.125 & 0.520 & 0.383 & 0.441 \\

\cmidrule(lr){1-10}

{Ours} & \textbf{0.451} & \textbf{0.310} & \textbf{0.219} & \textbf{0.159} 
       & \textbf{0.333} & \textbf{0.177}  & {0.505} & \textbf{0.656} & \textbf{0.571} \\

\bottomrule
\end{tabular}
\end{table}





The prompt for system is "You are a professional radiologists", followed by the user prompt: "Please provide a radiology report of the following chest X-ray image. Keep only the 'Findings' and 'Impression' sections in your report" and the CXR image. The findings section were then retrieved by extracting the part between "findings" and "impression" via simple text search.  

As shown in Tables~\ref{tab:iu_mimic_results}, our method achieved state-of-the-art performance on the MIMIC-CXR dataset across both {NLG and CE metrics.}
In particular, our model obtains notable improvements on NLP metrics, indicating stronger capability in modeling long-range dependencies.
Moreover, our approach yields substantial gains in  {and CE} metrics, demonstrating its effectiveness in generating clinically accurate reports.

\subsubsection{Ablation Study}
Table~\ref{tab:mimic_results} presents an ablation study on the MIMIC-CXR dataset to evaluate the individual and combined effects of the \blue{clinical reward} and group-relative training. \blue{We did not separately ablate the two $\lambda$-related designs because they are coupled: $\lambda$ conditioned decoder enables test-time control, while $\lambda$ weighted rewards provide the supervision during training that makes this control meaningful.} Without either component, the model achieves relatively limited performance on both NLG and CE metrics. Introducing the \blue{clinical reward} alone leads to substantial improvements in CE metrics, demonstrating its effectiveness in enhancing clinical correctness. Applying group-relative training alone also yields notable gains across both NLG and CE metrics. When both components are jointly applied, the model achieves the best overall performance across all metrics, highlighting their complementary roles in improving both linguistic quality and clinical efficacy.

\begin{table}[h]
\centering
\caption{Ablation study of \blue{clinical reward} and group-relative training on the MIMIC-CXR dataset.}
\label{tab:mimic_results}
\setlength{\tabcolsep}{4.5pt}
\scriptsize
\begin{tabular}{cc cccccc ccc}
\toprule
\multicolumn{2}{c}{Method} 
& \multicolumn{6}{c}{NLG Metrics} 
& \multicolumn{3}{c}{CE Metrics} \\ 
\cmidrule(lr){1-2} \cmidrule(lr){3-8} \cmidrule(lr){9-11}
Clinical & Group-rel. 
& B-1 & B-2 & B-3 & B-4 & RG-L & MET. 
& PREC. & REC. & F1 \\ 
\midrule

$\times$ & $\times$
& 0.400 & 0.253 & 0.171 & 0.120 & 0.276 & 0.154 
& 0.392 & 0.335 & 0.342 \\

$\checkmark$ & $\times$
& 0.404 & 0.260 & 0.180 & 0.129 & 0.312 & 0.153 & \textbf{0.590} & 0.330 & 0.423 \\

$\times$ & $\checkmark$
& 0.433 & 0.298 & 0.212 & 0.153 & \textbf{0.339} & 0.170 & 0.546 & 0.396 & 0.459 \\

$\checkmark$ & $\checkmark$
& \textbf{0.451} & \textbf{0.310} & \textbf{0.219} & \textbf{0.159} 
& {0.333} & \textbf{0.177}  & {0.505} & \textbf{0.656} & \textbf{0.571} \\

\bottomrule
\end{tabular}
\end{table}





\begin{figure*}[!t]
\centering

\begin{minipage}{0.29\textwidth}
\centering

\includegraphics[width=\linewidth]{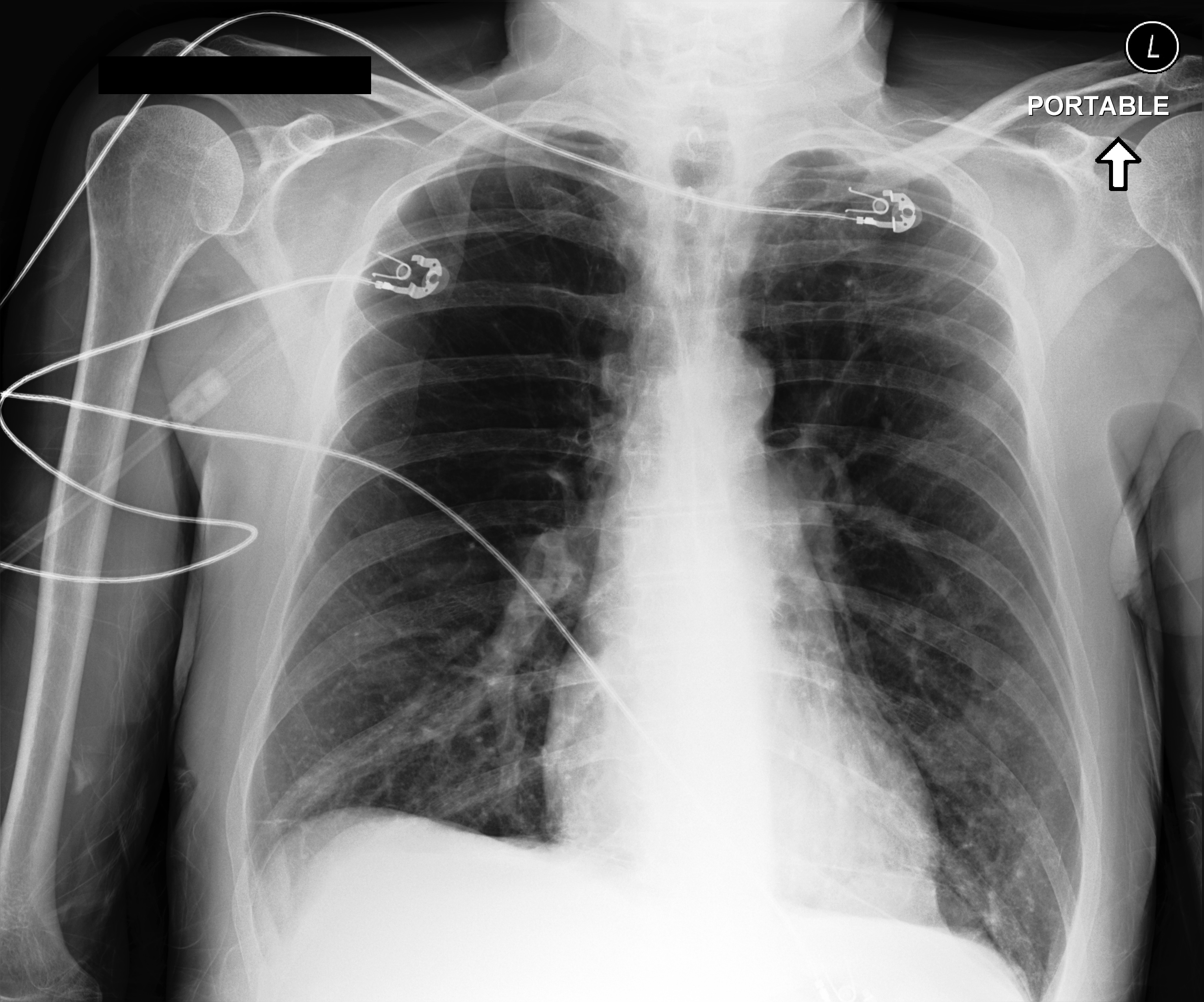}

\vspace{7mm}

\includegraphics[width=\linewidth]{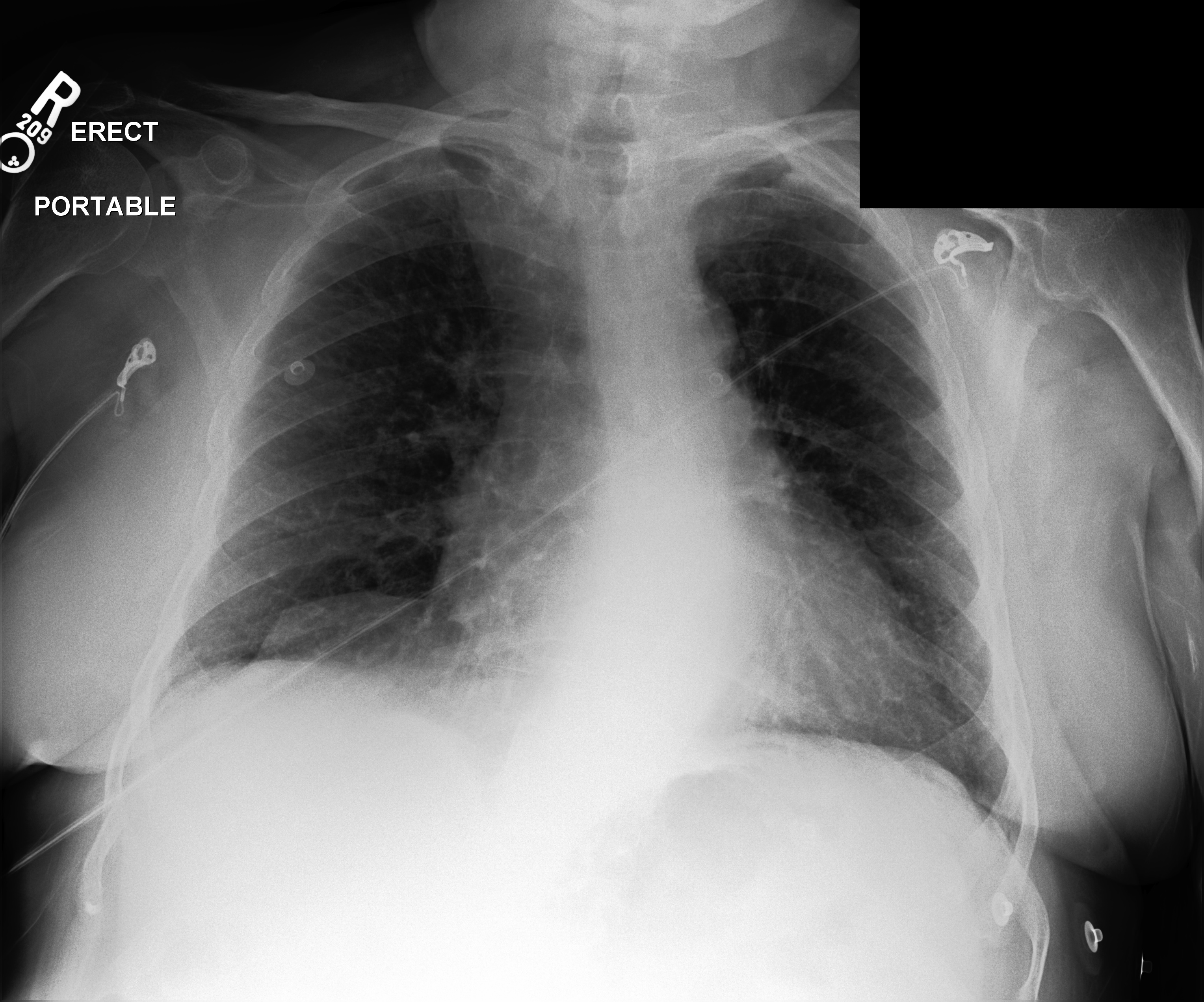}

\end{minipage}
\hfill
\begin{minipage}{0.7\textwidth}

\scriptsize

\fbox{\parbox{\linewidth}{
\textbf{Ground Truth:} ap portable upright chest radiograph was provided . the \blueee{lungs are hyperinflated} with upper lobe lucency compatible with emphysema . no focal consolidation effusion or pneumothorax seen . cardiomediastinal silhouette is normal . bony structures are intact .
}}

\vspace{1mm}

\fbox{\parbox{\linewidth}{
\textbf{Baseline:} ap portable upright view of the chest . overlying ekg leads are present . there is no focal consolidation effusion or pneumothorax . the cardiomediastinal silhouette is normal . imaged osseous structures are intact .
}}

\vspace{1mm}

\fbox{\parbox{\linewidth}{
\textbf{Ours:} ap view of the chest . \blueee{the lungs are hyperinflated .} the lungs are clear . there is no focal consolidation . no pleural effusion or pneumothorax . the cardiomediastinal silhouette is normal . the mediastinal and hilar contours are normal .
}}

\vspace{1.5mm}

\fbox{\parbox{\linewidth}{
\textbf{Ground Truth:} \blueee{cardiomegaly is stable} . there is no focal consolidation concerning for pneumonia . there is no pleural effusion pneumothorax or pulmonary edema . scoliosis is again noted . an old left clavicular deformity is noted .
}}

\vspace{1mm}

\fbox{\parbox{\linewidth}{
\textbf{Baseline:} as compared to the previous radiograph there is no relevant change . low lung volumes . \reddd{borderline size of the cardiac silhouette} without pulmonary edema . no pleural effusions . no pneumonia no pneumothorax .
}}

\vspace{1mm}

\fbox{\parbox{\linewidth}{
\textbf{Ours:} ap view of the chest . the lungs are clear . there is no focal consolidation . no pleural effusion or pneumothorax . \blueee{the cardiac silhouette is enlarged .} the mediastinal and hilar contours are normal .
}}

\end{minipage}

\caption{Qualitative comparison of radiology reports generated by the baseline method and our method with ground truth references. Main improvements are highlighted in blue.}
\label{fig:qualitative}

\end{figure*}

\subsubsection{Quantitative Analysis}
We further present qualitative analysis of representative samples in Fig. \ref{fig:qualitative} to illustrate the differences between our method and the baseline model  {\cite{chen2020generating}}. The reports generated by our method demonstrated improved clinical accuracy and closer alignment with the ground truth. In the first case, our model correctly identified the hyperinflated lung which was missed by the baseline model. In the second case our model confirmed the enlarged cardiac silhouette whereas the baseline model described the size as "borderline". These examples demonstrate that the reports generated by our framework are more consistent with the ground truth and better reflect clinically meaningful findings.

\section{Conclusion}

This paper proposed a novel approach to enhance clinical consistency and controllability in RRG. Our method enables flexible precision recall control through a hybrid natural language and clinical efficacy reward mechanism, allowing the model to better capture clinically relevant findings while maintaining semantic coherence. Extensive experiments on the MIMIC-CXR dataset demonstrated the effectiveness and robustness of our approach compared with existing methods. However, the current model is limited by the small size of the model as well as uncleaned training data, e.g., some data points were lateral CXR only, or the reports mentioned previous exam but the input had only the current CXR. In future works, we plan to extend \blue{Group reward-decoupled normalization (GDPO) \cite{liu2026gdpo} to our method and extend} the framework to LLMs such as MedGemma and finetune it with a cleaner dataset. We also plan to perform more comprehensive evaluations of the methods with more advanced metrics \cite{yu2023evaluating}, human reader study, and institutional data, \blue{and compare our method with recent approaches such as CURE \cite{messina2026cure} for curriculum-guided anatomy grounding.}

\subsubsection{\blue{Acknowledgments}}
\red{This work was supported in part by NIBIB under Award Number R01EB035394. The content is solely the responsibility of the authors and does not necessarily represent the official views of NIH. }

\subsubsection{\blue{Disclosure of Interests}}
\blue{The authors have no competing interests in the paper.}
    



%
%
%

\bibliographystyle{unsrt}
\bibliography{Paper-4076}
%




\end{document}